\documentclass[10pt,twocolumn,letterpaper]{article}

\usepackage{cvpr}              

\definecolor{cvprblue}{rgb}{0.21,0.49,0.74}
\usepackage[pagebackref,breaklinks,colorlinks,allcolors=cvprblue]{hyperref}
\usepackage{natbib}
\usepackage{colortbl}
\definecolor{lightgreen}{rgb}{0.84, 1.0, 0.84}
\usepackage{multirow}

\usepackage{pifont} 
\usepackage{booktabs} 
\usepackage{adjustbox}
\usepackage{arydshln}
\usepackage{tabularx}

\usepackage{pifont} 

\usepackage{tcolorbox}
\tcbuselibrary{breakable,skins} 

\newtcbox{\hlblue}{on line, arc=1pt, colback=blue!12,  boxrule=0pt, boxsep=1pt,
  left=2pt,right=2pt, top=1pt,bottom=1pt}
\newtcbox{\hlyellow}{on line, arc=1pt, colback=yellow!12, boxrule=0pt, boxsep=1pt,
  left=2pt,right=2pt, top=1pt,bottom=1pt}
\newtcbox{\hlgreen}{on line, arc=1pt, colback=green!12, boxrule=0pt, boxsep=1pt,
  left=2pt,right=2pt, top=1pt,bottom=1pt}
\newtcbox{\hlorange}{on line, arc=1pt, colback=orange!12, boxrule=0pt, boxsep=1pt,
  left=2pt,right=2pt, top=1pt,bottom=1pt}
\newtcbox{\hlpurple}{on line, arc=1pt, colback=purple!12, boxrule=0pt, boxsep=1pt,
  left=2pt,right=2pt, top=1pt,bottom=1pt}
\newtcolorbox{promptbox}{
  breakable,                  
  width=\linewidth,
  colback=gray!10,            
  colframe=gray!30,           
  arc=2mm,                    
  boxrule=0.3pt,              
  left=6pt,right=6pt,top=6pt,bottom=6pt
}

\usepackage{amsfonts}
\usepackage{amsmath}
\usepackage{xcolor} 
\usepackage[table]{xcolor}
\definecolor{lightgreen}{rgb}{0.84, 1.0, 0.84}
\usepackage{booktabs, multirow, makecell, xcolor}
\usepackage{siunitx}
\newcommand{\up}{\raisebox{0.6pt}{$\uparrow$}} 
\usepackage{arydshln}

\def\paperID{15295} 
\def\confName{CVPR}
\def\confYear{2025}

\title{MMD-Thinker: Adaptive Multi-Dimensional Thinking for Multimodal Misinformation Detection}

\author{
Junjie Wu$^1$ \ \  \  Guohong Fu$^{1,2}$\thanks{Corresponding author.}\\
$^1$School of Computer Science and Technology, Soochow University\\
$^2$Institute of Artificial Intelligence, Soochow University\\
{\tt\small 20224027010@stu.suda.edu.cn, ghfu@suda.edu.cn}
}

\begin{document}
\maketitle

\begin{abstract}
Multimodal misinformation floods on various social media, and continues to evolve in the era of AI-generated content (AIGC). The emerged misinformation with low creation cost and high deception poses significant threats to society. While recent studies leverage general-purpose multimodal large language models (MLLMs) to achieve remarkable results in detection, they encounter two critical limitations: (1) Insufficient reasoning, where general-purpose MLLMs often follow the uniform reasoning paradigm but generate inaccurate explanations and judgments, due to the lack of the task-specific knowledge of multimodal misinformation detection. (2) Reasoning biases, where a single thinking mode make detectors a suboptimal path for judgment, struggling to keep pace with the fast-growing and intricate multimodal misinformation. In this paper, we propose MMD-Thinker, a two-stage framework for multimodal misinformation detection through adaptive multi-dimensional thinking. First, we develop tailor-designed thinking mode for multimodal misinformation detection. Second, we adopt task-specific instruction tuning to inject the tailored thinking mode into general-purpose MLLMs. Third, we further leverage reinforcement learning strategy with a mixed advantage function, which incentivizes the reasoning capabilities in trajectories. Furthermore, we construct the multimodal misinformation reasoning (MMR) dataset, encompasses more than 8K image-text pairs with both reasoning processes and classification labels, to make progress in the relam of multimodal misinformation detection. Experimental results demonstrate that our proposed MMD-Thinker achieves state-of-the-art performance on both in-domain and out-of-domain benchmark datasets, while maintaining flexible inference and token usage. Code will be publicly available at Github.
\end{abstract}    
\section{Introduction}
\label{sec:intro}

\begin{figure}[t]
  \includegraphics[width=\columnwidth]{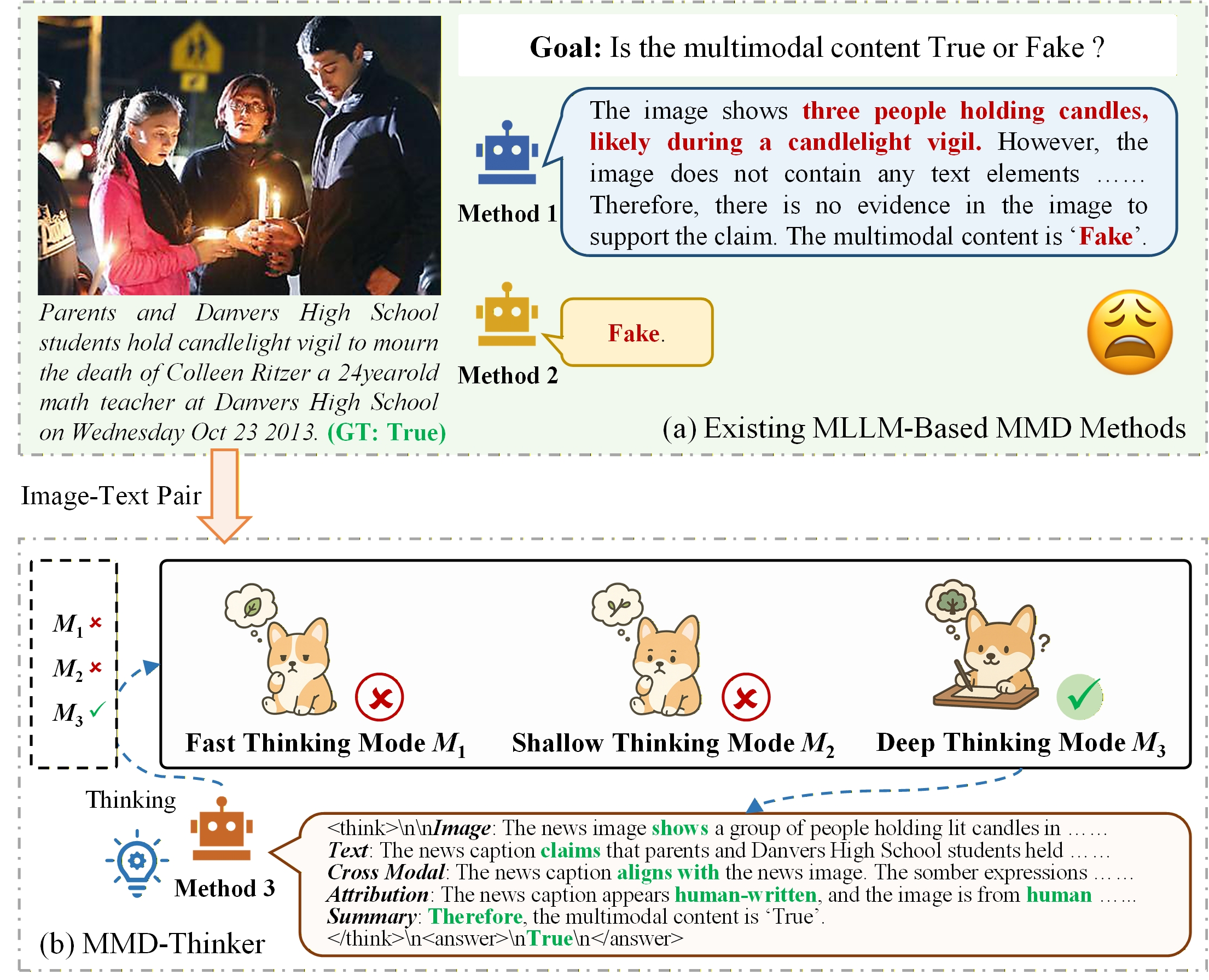}
  \caption{Illustrations of different methods in the task of multimodal misinformation detection.}
  \label{fig:1}
\end{figure}

The rapid expansion of AI-generated content (AIGC) technology in recent years has revolutionized how multimodal content is created, while significantly lowering the barrier for producing multimodal misinformation \citep{Wang_2025_CVPR, yan-etal-2025-trust}. As a results, a vast amount of multimodal misinformation has flooded social media platforms, such as Twitter and Facebook, posing unprecedented threats to societal stability \citep{ma-etal-2024-event, Zhang_Zhang_Zhou_Huang_Li_2024}. Although professional fact-checkers work on checking the authenticity of multimodal content, they fail to keep pace with the fast-growing misinformation. Hence, automated methods have become an urgent need to effectively detect multimodal misinformation in the AIGC era.

In the meantime, Multimodal Large Language Models (MLLMs) such as Qwen2.5-VL \citep{qwen2.5-VL} have presented promising progress in visual question answering \citep{Fang_2025_CVPR, shah-etal-2025-analyzing}, mathematical reasoning \citep{liu2025noisyrollout, yu-etal-2025-self-error} and code generation tasks \citep{luo-etal-2025-tree, guo-etal-2025-personality}. These remarkable language understanding and reasoning capabilities make MLLMs an effective solution for enhancing multimodal misinformation detection.

Recently, a series of studies \citep{shao2024deepseekmathpushinglimitsmathematical, NEURIPS2024_c7f43ada, 10.1145/3701716.3715599} attempt to mimic fast thinking by leveraging the task-specific instruction tuning, as shown in Figure \ref{fig:1}. Such detectors make quick decisions without deliberate thinking process within a small token budget. However, this depends on heuristics from vast data patterns, resulting in their cognitive biases \citep{zhang2025othinkr1intrinsicfastslowthinking}. In contrast, the detectors with slow thinking employ a thorough thinking processes before decision-making, as shown in Figure \ref{fig:1}. However, the characteristic of the uniform reasoning paradigm is insufficient for debunking multimodal misinformation. Moreover, a single-thinking mode limits the effectiveness of detectors in enormous and complex misinformation. Therefore, it is critical to empower detectors with dedicated reasoning in coping with multimodal misinformation in this new era of AIGC.

To reach the target above, we propose MMD-Thinker, the first detection framework for multimodal misinformation detection through adaptive multi-dimensional thinking. The MMD-Thinker primarily consists of three modules: (1) The \textit{multi-dimensional thinking mode design} module includes a spectrum from quick response and shallow thinking to deep deliberation. These modes align with fact-checker's cognitive behavior during reasoning. We conduct a comprehensive experiment to illustrate the effectiveness and efficiency of such design for discriminating between fake and real. (2) The \textit{multi-dimensional thinking mode learning} module guides baseline models to establish basic reasoning capabilities for detection by leveraging task-specific instruction tuning. (3) The \textit{adaptive multi-dimensional thinking mode policy optimization} module utilizes group-relative policy optimization with a mixed advantage function from the perspectives of both mode and sample, thereby enabling the model to actively incentivize the reasoning capabilities in trajectory. This framework effectively improves flexible inference and token usage in multimodal misinformation detection.

Additionally, to unlock the reasoning capabilities of existing open-source MLLMs in the multimodal misinformation domain, we introduce the multimodal misinformation reasoning (MMR) dataset, a novel resource designed for task-specific instruction tuning with the injection of thinking modes. MMR consists of more than 8K annotated samples. Each sample consists of an image-text pair with both reasoning process and judgment. It serves as a valuable resource for both training and evaluation of advancing MLLM-based multimodal misinformation detection. Extensive experiments conducted on in-domain and out-of-domain benchmark datasets demonstrate that MMD-Thinker achieves state-of-the-art results across different evaluation metrics, emphasizing its effectiveness for multimodal misinformation detection in the new era of AIGC. Our main contributions are three-fold:

\begin{itemize}
    \item We propose MMD-Thinker, the first adaptive multi-dimensional thinking framework for multimodal misinformation detection. It consists of thinking design, thinking learning, and thinking policy optimization. MMD-Thinker establishes a novel paradigm for multimodal misinformation detection by dynamically employing thinking mode and accurately generating answer.
    \item We construct MMR, the expert-annotated multimodal misinformation dataset, including 8K+ image-text pairs with reasoning processes and classification labels. MMR enables multimodal misinformation reasoning beyond discrete misinformation detection.
    \item We conduct experiments on both in-domain and out-of-domain benchmark dataset demonstrate that the MMD-Thinker achieves the significant improvements compared to general-purpose MLLMs across different experimental settings, \textit{e.g.,} zero-shot, task-specific instruction tuning, and reinforcement learning. 
\end{itemize}
\section{Related Work}
\label{sec:Related Work}

\begin{figure*}[t]
  \includegraphics[width=\textwidth]{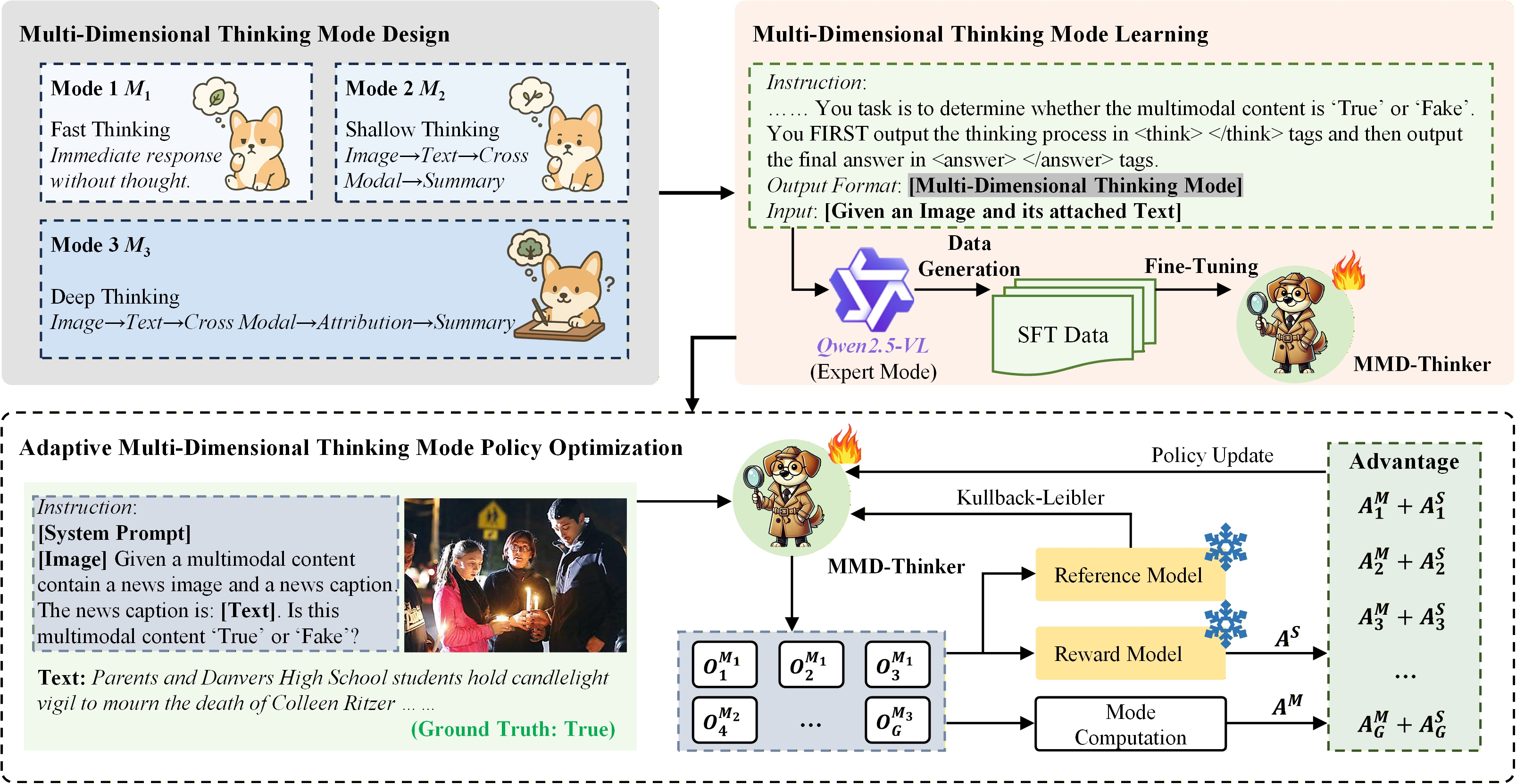}
  \caption{Overview of the MMD-Thinker framework for multimodal misinformation detection. The framework consists of three critical modules: Multi-dimensional thinking mode design, multi-dimensional thinking mode learning, and adaptive multi-dimensional thinking mode policy optimization.}
  \label{fig:2}
\end{figure*}

\subsection{MLLMs and Multimodal Misinformation Detection}

With the rapid advancement of general-purpose multimodal large language models (MLLMs) \citep{pmlr-v202-li23q, NEURIPS2023_6dcf277e, yao2024minicpmvgpt4vlevelmllm, chen2023minigptv2largelanguagemodel, qwen2.5-VL}, the increasing attention of researchers has focused on leveraging the inherent capabilities of MLLMs for different domain applications \citep{NEURIPS2024_c7f43ada, zhang-etal-2023-video, lei2024instructercreformingemotionrecognition}. In the realm of multimodal misinformation detection, existing approaches with general-purpose MLLMs primarily focus on two orientations, including zero-shot and instruction tuning. Recent works \citep{liu2024mmfakebench, wang2024mfcbenchbenchmarkingmultimodalfactchecking} use standard prompting with the direct-answer template to predict classification labels. However, even advanced closed-source models likewise GTP-4V \citep{jiang2025largevisuallanguagemodelsgood} fail short in achieving accurate predictions, due to the lack of task-specific instruction tuning on multimodal misinformation dataset. Subsequently, researchers have begun instruction-tuned MLLMs on instruction-following datasets \citep{huang2025emotionqwenunifiedframeworkemotion, 10.1007/978-3-031-57916-5_8, lei2024instructercreformingemotionrecognition}. Evaluations of these models show the improvements in detection performance, they struggle to provide explanations or infer forgery generation trajectories for judgments \citep{zhang2025factr1explainablevideomisinformation}. \textit{In contrast, our proposed MMD-Thinker employs reinforcement learning to optimize the trajectory, thus enhancing detection and reasoning capabilities.}

\subsection{Reasoning in MLLMs}

Current approaches proposed to solve the lack of reasoning processes in the task of classification can be three fronts. Early work employs standard prompting with the reasoning template to achieve both reasoning process and judgment \citep{Xu_Sun_Zhai_Li_Liang_Li_Du_2025, Xu_2025_CVPR, 10.1007/978-3-031-78456-9_25}. However, they often lack domain-specific knowledge, resulting in the limitation of the reasoning capabilities. Some studies extend general-purpose MLLMs, such as Qwen2.5-VL \citep{qwen2025qwen25technicalreport}, to visual and multimodal tasks by constructing supervised data that contains both reasoning processes and judgments \citep{guo-etal-2025-mammoth, Tong_2025_ICCV}. However, these instruction-tuned models often follow a certain predefined reasoning style while ignoring potentially reasoning paths. \textit{In this work, we inspired by hierarchical cognitive control theory \citep{KOECHLIN2007229} devise multiple thinking modes for multimodal misinformation detection, and further make detectors dynamically reasoning for judgment.}
\section{Methodology}
\label{sec:Methodology}

In this section, we introduce our proposed MMD-Thinker framework, as shown in Figure \ref{fig:2}. Sec. \ref{sec:3.1} introduces the multi-dimensional thinking mode design module, where the designed thinking mode covers a spectrum from quick response and shallow thinking to deep deliberation by aligning with fact-checker’s cognitive behavior. Sec. \ref{sec:3.2} elaborates the multi-dimensional thinking mode learning module for conducting the injection of thinking mode through multimodal misinformation detection fune-tuning. Sec. \ref{sec:3.3} describes the adaptive multi-dimensional thinking mode policy optimization module as the core component of the MMD-Thinker, where we adopt a mixed advantage function that includes both mode and sample, establishing a flexible and efficient reasoning process for judgment.


\subsection{Multi-Dimensional Thinking Mode Design}
\label{sec:3.1}

To align with human cognitive behavior during the process of multimodal misinformation detection, we consider the fact that cognitive control is presented in a hierarchical structure, including the reactive level, semantic level, and prospective level \citep{KOECHLIN2007229, BADRE2008193, wang2025ampo, zhu2025conciseadaptivethinkinglarge}. As a result, we propose a multi-dimensional thinking mode design module with three different hierarchical levels. This works on the management of both goals and actions at distinct levels. Specifically, the designed thinking mode is presented as follows:

(1) Quick response: A fast thinking serves as the basic linguistic mode at the reactive level. This thinking mode consists of a decision (``real'' or ``fake'') without any reasoning process.

(2) Semantic analysis: A shallow thinking refers to the basic interaction mode at the semantic level. This thinking mode emphasizes on understanding the semantics of intra- and inter-modality within image-text pairs. It consists of judgment and reasoning process that contains a series of actions, \textit{i.e.,} \textit{image analysis}, \textit{text analysis}, \textit{cross-modal analysis}, and \textit{summary}. \textit{Image analysis} is to analyze the key elements and content of image. \textit{Text analysis} aims to describe the details of text. \textit{Cross-modal analysis} ensures the consistencies between image and text. \textit{Summary} provides a conclusion for the final classification label.

(3) Prospective simulation: A deep deliberation indicates the advanced simulation mode at the prospective level. It based on semantic analysis further introduces an action \textit{attribution} for discriminating between AI-generated and human-generated. This thinking mode facilitates detectors to simulate different types of multimodal misinformation, thereby performing deeper thinking for more complex and challenge scenarios.

\begin{table*}[t]
\centering
\resizebox{\textwidth}{!}{
\begin{tabular}{lr
r r r r
r r r r
c c}
\toprule
\multirow{3}{*}{Methods} &
\multirow{3}{*}{Data Size} &
\multicolumn{4}{c}{\multirow{2}{*}{In-Domain}} &
\multicolumn{4}{c}{Out-of-Domain} &
\multirow{3}{*}{\textbf{Avg. Tokens}} &
\multirow{3}{*}{\textbf{Avg. F1 \up}} \\
& & & & & &
\multicolumn{2}{c}{PHEME} &
\multicolumn{2}{c}{Twitter} & & \\
\cmidrule(lr){3-6} \cmidrule(lr){7-10}
& &
\textbf{Acc. \up} & \textbf{F1 \up} & \textbf{Pre. \up} & \textbf{Rec. \up} &
\textbf{Acc. \up} & \textbf{F1 \up} &
\textbf{Acc. \up} & \textbf{F1 \up} & & \\
\midrule
\multicolumn{12}{l}{\textbf{\textit{\textcolor{gray}{General-purpose MLLMs}}}}\\
GPT-5-Mini                    &            & 80.48 & 73.87 & 75.32 & 72.84 & 44.67 & 44.07 & 57.05 & 54.48 &  99 & 57.47 \\
Gemma 3-12B                   &            & 68.34 & 67.54 & 72.23 & 77.63 & 34.88 & 33.30 & \textbf{61.93} & 61.35 & 116 & 54.06 \\
Llama-3.2-11B-Vision          &            & 63.95 & 63.39 & 66.27 & 68.49 & 42.31 & 41.76 & 57.72 & 57.65 & 166 & 54.27 \\
Qwen2-VL-7B                   &            & 73.80 & 71.18 & 70.97 & 76.20 & 41.34 & 40.57 & 61.41 & \textbf{61.41} &  87 & 57.72 \\
Qwen2.5-VL-7B        &            & \textbf{82.10} & \textbf{78.46} & \textbf{77.34} & \textbf{80.20} & 43.76 & 39.74 & 56.06 & 55.06 & 124 & 57.75 \\
Qwen2.5-VL-3B                 &            & 75.30 & 69.47 & 68.98 & 70.12 & \textbf{51.49} & \textbf{44.27} & 57.67 & 56.34 & \textbf{86} & \textbf{59.10} \\
\midrule
\multicolumn{12}{l}{\textbf{\textit{\textcolor{gray}{Misinformation Detectors}}}}\\
\multicolumn{12}{l}{Qwen2.5-VL-3B}\\
\quad w/ SFT                   & \textcolor{cyan!90}{6K}        & 88.60 & 85.15 & 86.00 & 84.41 & 51.16 & 44.63 & 59.36 & 54.54 & 220 & 61.44 \\
\quad w/ SFT                   & \textcolor{cyan!90}{7K}        & 89.19 & 86.07 & 86.52 & 85.64 & 51.65 & 44.08 & 61.04 & 56.63 & 229 & 62.26 \\
\quad w/ SFT + Vanilla GRPO    & \textcolor{cyan!90}{6K}+\textcolor{red!90}{1K}     & 89.90 & 87.11 & 87.16 & 87.07 & 51.32 & 45.57 & 61.41 & 57.83 & 212 & 63.50 \\
\rowcolor{lightgreen}
\quad w/ SFT + MMPO (\textit{Ours}) & \textcolor{cyan!90}{6K}+\textcolor{red!90}{1K} & \textbf{90.70} & \textbf{88.40} & \textbf{87.65} & \textbf{89.27} & \textbf{52.31} & \textbf{46.80} & \textbf{62.84} & \textbf{59.12} & \textbf{192} & \textbf{64.77} \\
\hdashline
\multicolumn{12}{l}{Qwen2.5-VL-7B}\\
\quad w/ SFT                   & \textcolor{cyan!90}{6K}        & 90.30 & 87.28 & 88.48 & 86.28 & 52.15 & 45.89 & 65.41 & 61.70 & 223 & 64.96 \\
\quad w/ SFT                   & \textcolor{cyan!90}{7K}        & 90.59 & 87.90 & 88.28 & 87.54 & 51.32 & 44.01 & 63.65 & 59.15 & 225 & 63.69 \\
\quad w/ SFT + Vanilla GRPO    & \textcolor{cyan!90}{6K}+\textcolor{red!90}{1K}     & 92.19 & 89.89 & 90.61 & 89.23 & 53.96 & 47.88 & 65.13 & 59.45 & 210 & 65.74  \\
\rowcolor{lightgreen}
\quad w/ SFT + MMPO (\textit{Ours}) & \textcolor{cyan!90}{6K}+\textcolor{red!90}{1K} & \textbf{92.90} & \textbf{90.74} & \textbf{91.78} & \textbf{89.83} & \textbf{58.25} & \textbf{50.86} & \textbf{68.60} & \textbf{62.53} & \textbf{175} & \textbf{68.04} \\
\bottomrule
\end{tabular}
} 
\caption{Performance comparison between our MMD-Thinker (SFT+MMPO) and other methods across in-domain and out-of-domain datasets. Best results are in bold.}
\label{tab:1}
\end{table*}

\subsection{Multi-Dimensional Thinking Mode Learning}
\label{sec:3.2}

General-purpose MLLMs often lack the domain-specific knowledge of multimodal misinformation detection, preventing their capabilities from establishing the relationships between reasoning processes and judgments. This results in the limitation of inherent capabilities of general-purpose MLLMs within reasoning paths. To cope with this issue, we adopt instruction tuning to extend the general-purpose MLLM Qwen2.5-VL \citep{qwen2.5-VL} to the task of multimodal misinformation detection for both reasoning process and judgment. Specifically, we follow the designed multi-dimensional thinking mode to construct a minimal and high-quality multimodal misinformation instruction-following dataset for end-to-end training. Based on the constructed dataset $\mathcal{D}_1$, the objective of this stage can be formulated as follows:
\begin{equation}
\mathcal{L}_1
= - \mathbb{E}_{(x,y)\sim \mathcal{D}_1}
\Bigg[
  \sum_{t=1}^{|y|}
  \log \pi_{\theta}( y_t \mid x, y_{<t} )
\Bigg],
\end{equation}

This task-specific instruction tuning makes the model adhere to the designed thinking mode during the process of detection. The details for constructing the multimodal misinformation instruction-following dataset can be found in the next version.


\subsection{Adaptive Multi-Dimensional Thinking Mode Policy Optimization}
\label{sec:3.3}

Although the general-purpose MLLM has access to the initial reasoning capabilities after the stage of instruction tuning, it faces the dilemma related to reasoning biases that limit their flexibility and effectiveness across different scenarios. To address the limitation above, we leverage reinforcement learning (RL) to further enhance the reasoning capabilities of our method. Specifically, we employ group-relative policy optimization (GRPO) \cite{shao2024deepseekmathpushinglimitsmathematical, wang2025ampo} in the RL algorithm for end-to-end training. Formally, the objective function of GRPO is calculated as:
\begin{equation}
\begin{aligned}
\mathcal{L}_2 
= -\mathbb{E}_{(x,y)\sim\mathcal{D}_2} \Bigg[
\frac{1}{G}\sum_{i=1}^{G} \Big(
\min\!\Big(
    \frac{\pi_\theta(y_i)}{\pi_{\text{old}}(y_i)} A_i,\,\\
    \operatorname{clip}\!\Big(
        \frac{\pi_\theta(y_i)}{\pi_{\text{old}}(y_i)},
        1-\epsilon,\,
        1+\epsilon
    \Big) A_i
\Big)
\\
\qquad\qquad
- \beta\, \mathcal{D}_{\mathrm{KL}}\!\left(\pi_\theta \,\|\, \pi_{\mathrm{ref}}\right)
\Big)
\Bigg],
\end{aligned}
\label{eq:grpo}
\end{equation}
where $\mathcal{D}_2$ denotes the training set. $\pi_\theta$, $\pi_{\text{old}}$, and $\pi_{\mathrm{ref}}$ denote the current policy model, the sampling policy model, and the fixed reference model, respectively. $G$ denotes the total number of responses. $\epsilon$ denotes the hyperparameter that controls the clipping range. $\beta$ denotes the hyperparameter for controlling the KL penalty strength. $\mathcal{D}_{\mathrm{KL}}\!\left(\pi_\theta \,\|\, \pi_{\mathrm{ref}}\right)$ indicates the KL divergence constraint. $A$ represents the mixed advantages for each response $y$ in terms of both mode and sample:
\begin{equation}
A_i = A^M_i + A^S_i,
\end{equation}
where $A^S$ represents the normalized advantage at the sample-level:
\begin{equation}
A^S_i =
\frac{R^s_i - \operatorname{mean}\big(\{R^s_1,\ldots,R^s_G\}\big)}
     {\operatorname{std}\big(\{R^s_1,\ldots,R^s_G\}\big)},
\label{eq:adv_le}
\end{equation}
where, $\operatorname{mean}$ and $\operatorname{std}$ denote the standard deviation and the arithmetic mean, respectively. $r^s_i$ represents the reward value of the $i$-th response from the policy model. It integrates detection accuracy reward $r_{acc}$ with reasoning format reward $r_{format}$. The former ensures the consistency between the ground truth and the final classification label. The latter enables the model to follow the standard structure of reasoning process. The reasoning process is fixed in ``<think> </think>'', and the final classification label is set in ``<answer> </answer>''.

The advantage $A^M$ is computed at the mode-level:
\begin{equation}
A^M_i =
\frac{R^m_i - \operatorname{mean}\big(\{R^{m(1)},\ldots,R^{m(n)}\}\big)}
     {\operatorname{std}\big(\{R^{m(1)},\ldots,R^{m(n)}\}\big)},
\label{eq:adv_mo}
\end{equation}
where, $n\in\{1,2,3\}$ denotes the number of the thinking mode. $R^m$ denotes the average reward within a certain thinking mode. The mixed advantage function can facilitate the model to adaptively select an appropriate thinking mode for providing the final judgment.

In summary, compared to the multi-dimensional thinking mode learning with task-specific instruction tuning that adheres to the basic thinking mode for multimodal misinformation detection, adaptive multi-dimensional thinking mode policy optimization emphasizes flexible inference and token usage across different scenarios.


\section{Experiment}
\label{sec:Experiment}

In this section, we conduct extensive experiments on various multimodal misinformation benchmark datasets, such as MMR, PHEME \citep{10.1007/978-3-319-67217-5_8} and Twitter \citep{13292}, to comprehensively evaluate the performance of our proposed MMD-Thinker.




\noindent \textbf{Implementation Detail.} All experiments are conducted using PyTorch \citep{NEURIPS2019_bdbca288} on 2$\times$NVIDIA A800 GPUs (80G), with both Qwen2.5-VL-3B and Qwen2.5-VL-7B as baseline models \citep{qwen2.5-VL}. In the stage of multi-dimensional thinking mode learning, we train for 3 epochs on 6K image-text pairs with a batch size of 8 and a learning rate of $2\times10^{-5}$. In the stage of adaptive multi-dimensional thinking mode policy optimization, we train for 8 epochs on 1K image-text pairs with a batch size of 2 and 8 samples per group.


\subsection{Results and Analyses}

In Table \ref{tab:1}, we can achieve the following observations: (1) Our proposed MMD-Thinker with Qwen2.5-VL-7B significantly surpasses all compared methods on both in-domain and out-of-domain benchmark datasets, resulting in more than a 7\% improvement in average F1-score. This strong performance demonstrates the efficacy of the MMD-Thinker and its potential to advance multimodal misinformation detection and reasoning capabilities. (2) Although general-purpose MLLMs have powerful understanding and generation capabilities in visual and multimodal tasks, they lack the extensive knowledge in the news domain. This limits their performance, even GPT-5-Mini. (3) Compared with SFT and vanilla GRPO, MMD-Thinker further achieves performance gains. The above results demonstrate the efficiency of MMD-Thinker in the task of multimodal misinformation detection.
\section{Conclusion}
\label{sec:Conclusion}

In this paper, we address the limitations of multimodal misinformation detection in the new era of AIGC. We propose MMD-Thinker, consisting of thinking design, thinking learning, and thinking optimization. Moreover, we construct the MMR dataset with structured reasoning processes that mimic fact-checkers. Extensive experiments show that the MMD-Thinker achieves state-of-the-art results across both in-domain and out-of-domain datasets.
{
    \small
    \bibliographystyle{ieeenat_fullname}
    \bibliography{main}
}


\end{document}